\pdfoutput=1

\documentclass[11pt]{article}

\usepackage[preprint]{acl}

\usepackage{times}
\usepackage{latexsym}

\usepackage[T1]{fontenc}

\usepackage[utf8]{inputenc}

\usepackage{microtype}

\usepackage{inconsolata}

\usepackage{graphicx}

\usepackage[T1]{fontenc}    
\usepackage{hyperref}       
\usepackage{url}            
\usepackage{booktabs}       
\usepackage{amsfonts}       
\usepackage{nicefrac}       
\usepackage{microtype}      
\usepackage{xcolor}         

\usepackage{inconsolata}
\usepackage{multicol}
\usepackage{float}
\usepackage{subcaption} 
\usepackage{caption}

\usepackage{tabularx}

\usepackage{graphicx}
\usepackage{booktabs}
\usepackage{array}
\usepackage{multirow}
\usepackage{amsmath}

\usepackage{paracol}
\usepackage{listings}
\usepackage{wrapfig}

\definecolor{darkgreen}{RGB}{0,100,0}

\title{Say Less, Mean More: \\Leveraging Pragmatics in Retrieval-Augmented Generation}

\author{%
Haris Riaz$^{*}$ \and Ellen Riloff \and Mihai Surdeanu\\
University of Arizona, Tucson, AZ, USA\\
\texttt{\{hriaz,msurdeanu\}@arizona.edu} \quad \texttt{riloff@cs.arizona.edu}
} 

\begin{document}
\maketitle

\def\thefootnote{*}\footnotetext{Corresponding author.}

\def\thefootnote{\arabic{footnote}}

\begin{abstract}
We propose a simple, unsupervised method that injects pragmatic principles in retrieval-augmented generation (RAG) frameworks such as Dense Passage Retrieval~\cite{karpukhin2020densepassageretrievalopendomain} to enhance the utility of retrieved contexts. 
Our approach first identifies which sentences in a pool of documents retrieved by RAG are most relevant to the question at hand, cover all the topics addressed in the input question and no more, and then highlights these sentences within their context, before they are provided to the LLM, without truncating or altering the context in any other way. We show that this simple idea brings consistent improvements in experiments on three question answering tasks (ARC-Challenge, PubHealth and PopQA) using five different LLMs. It notably enhances relative accuracy by up to 19.7\% on PubHealth and 10\% on ARC-Challenge compared to a conventional RAG system.
\end{abstract}

\section{Introduction}

\label{sec:intro}
Retrieval-augmented generation (RAG)~\cite{lewis2020retrieval} has emerged as a solution to the limited knowledge horizon of large language models (LLMs). RAG combines ``pre-trained parametric and non-parametric memory for language generation,''~\cite{lewis2020retrieval} with the non-parametric memory typically retrieved from large collections of documents. RAG has been shown to dramatically improve the performance of LLMs on various question-answering and reasoning tasks (see section \ref{sec:related-work}). However, we argue that RAG often overwhelms the LLM with too much information, only some of which may be relevant to the task at hand. This contradicts Grice's four maxims of effective communication~\cite{grice1975logic}, which state that the information provided should be ``as much as needed, and no more'' and that it should be ``as clear, as brief'' as possible. The four maxims are enumerated as follows: (1) \textit{Maxim of Quantity}: Provide as much information as needed, but no more; (2) \textit{Maxim of Quality}: Be truthful; avoid giving information that is false or unsupported; (3) \textit{Maxim of Relation}: Be relevant, sharing only information pertinent to the discussion; (4) \textit{Maxim of Manner}: Be clear, brief, and orderly; avoid obscurity and ambiguity. While these maxims were originally formulated in the context of human communication, we argue that they are also applicable in a RAG setting.

We propose a simple, unsupervised method that injects pragmatics in any RAG framework\footnote{Code is available at: \url{https://github.com/hriaz17/SayLessRAG}}. In particular, our method: (a) identifies which sentences in a pool of documents retrieved by RAG are most relevant to the question at hand (maxim of relation), and cover all the topics addressed in the input question and no more (maxim of quantity and manner);\footnote{We envision that the maxim of quality could be considered too by identifying factual statements~\cite{rudinger-etal-2018-neural-models}. We leave this for future work.} and (b) highlights these sentences within their original contexts before they are provided to the LLM. Table~\ref{tab:example} shows an example of our method in action. 

The contributions of our paper are:
{\flushleft {\bf (1)}} We introduce a strategy to introduce pragmatics into any RAG method such as Dense Passage Retrieval~\cite{karpukhin2020densepassageretrievalopendomain}. To our knowledge, we are the first to investigate the impact of pragmatics for RAG.
\vspace{-2mm}
{\flushleft {\bf (2)}} We evaluate the contributions of pragmatics in RAG on three datasets: ARC-Challenge \cite{Clark2018ThinkYH}, PubHealth \cite{kotonya2020explainable} and PopQA \cite{mallen2023llm_memorization} and with five different LLMs ranging from 1B to 7B parameters: Mistral-7B-Instruct-v0.1 \cite{jiang2023mistral7b}, Alpaca-7B \cite{alpaca}, Llama2-7B-chat \cite{touvron2023llama2openfoundation}, Qwen2.5-3B \cite{qwen2.5} and AMD-OLMo-1B-SFT \cite{AMD-OLMo}. Our results indicate that pragmatics helps the most when the QA task primarily involves single-hop or multi-hop logical deduction where the highlighted evidence comprises factual statements that can be sequentially chained to derive the answer. Our post-hoc analysis further shows that this approach fares especially well for queries that benefit from analogical reasoning; with highlighted evidence sentences resembling in-context learning exemplars, proving especially useful for smaller language models with limited reasoning capabilities such as AMD-OLMo-1B-SFT, enabling a 10\% relative improvement on ARC-Challenge for this model.
{\flushleft {\bf (3)}} We find that pragmatics is less effective when the QA task requires arithmetic manipulation, or involves subtleties such as \textit{double negation}. Furthermore, we find that for factoid QA tasks, if a set of ambiguous contexts are first retrieved by DPR for a given query where 
the query lacks disambiguating information and multiple plausible answers could be derived, our method struggles to identify the appropriate evidence sentences for highlighting. In such cases, incorrect evidence highlighting can yield a slight degradation in LLM performance.
{\flushleft {\bf (4)}} Our empirical evidence suggests that our method is complementary when paired with a strong retriever like DPR; in favorable cases it can improve performance by up to 20\%, while exhibiting minimal degradation (approximately 1\%) in less optimal scenarios. Thus, we present it as a low risk and low overhead default augmentation to standard DPR implementations.

\begin{table}
\begin{small}
\begin{tabular}{>{\centering\arraybackslash}m{3mm}|p{0.4\textwidth}}
  \toprule
  \rotatebox[origin=b]{90}{
  Highlighted evidence
  ~~~~~~~} &
  \vspace{-9mm}
  [\dots] 
  Bats are famous for using echolocation to hunt down their prey, using sonar sounds to capture them in the dark. Another reason for nocturnality is avoiding the heat of the day. \textcolor{blue}{{\bf <evidence>This is especially true in arid biomes like deserts, where nocturnal behavior prevents creatures from losing precious water during the hot, dry daytime.</evidence>}} This is an adaptation that enhances osmoregulation. One of the reasons that (cathemeral) lions prefer to hunt at night is to conserve water.
  \\
  \midrule
  \rotatebox[origin=c]{90}{MCQ~~~~~~~~~~~~~~~~~~~~~} & 
  \vspace{-7mm}
  Question: Many desert animals are only active at night. How does being active only at night most help them survive in a hot desert climate?
  \vspace{2mm}
  
  Choices: 
  \vspace{-2mm}
  \begin{enumerate}
    \renewcommand{\labelenumi}{\Alph{enumi}.}
    \item They can see insects that light up at night.
    \vspace{-2mm}
    \item Their bodies lose less water in the cool night air.
    \vspace{-2mm}
    \item They are able to find more plant food by moonlight.
    \vspace{-2mm}
    \item Their bodies absorb sunlight in the daytime while they sleep.
  \end{enumerate} \\
  \bottomrule
\end{tabular}
\caption{\footnotesize Example of a multiple-choice question (MCQ) from the ARC-C dataset \cite{Clark2018ThinkYH} together with a fragment of a supporting document retrieved, in which the relevant evidence is highlighted with ``$<$evidence$>$'' tokens by our pragmatics-inspired algorithm. This evidence highlighting allows the downstream LLM to identify the correct answer (option B).}
\label{tab:example}
\end{small}
\end{table}

\section{Related Work}
\label{sec:related-work}
Since it was first proposed~\cite{lewis2020retrieval}, RAG has become an essential arrow in the quiver of LLM tools. However, many of the proposed RAG approaches rely on supervised learning to jointly optimize the retrieval component and the LLM~\citep[inter alia]{lewis2020retrieval,guu2020realmretrievalaugmentedlanguagemodel,xu2024recomp,kim2024reragimprovingopendomainqa} or to decide ``when to retrieve''~\cite{asai2024selfrag}. Instead, our approach is training free: it uses a set of unsupervised heuristics that approximate Grice's maxims (refer to Section~\ref{sec:intro}). 
Part of our method is similar to Active-RAG, which also reformulates the input query \cite{jiang2023activeretrievalaugmentedgeneration}. However, unlike Active-RAG, we use pragmatics to reformulate the input query and retrieve evidence for it,  instead of relying on LLM probabilities.
Our work is also similar to~\cite{xu2024recomp} and \cite{sarthi2024raptor}, which also touch on pragmatics by reducing the quantity of text presented to the LLM through summarization. However, the method used in \cite{xu2024recomp} is supervised. Furthermore, both of these methods exhibit considerably higher overhead compared to our proposed approach, which relies on simple yet robust heuristics. 

Our method adopts a {\em pre-retrieval} reasoning approach that is complementary to post-retrieval reasoning approaches such as \cite{trivedi2023interleavingretrievalchainofthoughtreasoning,kim2023treeclarificationsansweringambiguous},
which reason after document retrieval.
Further, we do not focus on reasoning about whether the retrieval was useful or not \cite{islam2024openrag}.
For example, current approaches that incorporate reasoning into the QA task, such as rStar \cite{qi2024mutualreasoningmakessmaller}, use an LLM to guide MCTS, where each intermediate step in the tree is verified by another LLM. \cite{jiang2024ragstarenhancingdeliberativereasoning} demonstrate that, rather than relying solely on the LLM’s parametric knowledge, retrieved contexts can also enhance tree search.
Another reasoning-based approach, STaR \cite{zelikman2022starbootstrappingreasoningreasoning}, employs an LLM to iteratively generate and refine a training set of rationales. The LLM is then fine-tuned on these rationales, generates a new set of rationales, and repeats the process.
In contrast, our method integrates reasoning directly into retrieval in a more efficient manner; specifically, we first reason about the task and then retrieve using the simple technique described in \cite{zheng2024stepbackevokingreasoning}.

Lastly, our work focuses on improving the utility of retrieved documents, somewhat similar to CRAG \cite{yan2024corrective}. However, we do not improve utility by retrieving more documents (e.g., from a web search) but rather by highlighting useful information already present in the current set of documents through pragmatics. Several previous methods, especially those based on summarization \cite{xu2024recomp} reduce the text by chopping it. Ours does not. The key idea of our work is to extract more utility \textit{while keeping the full text}.
\begin{figure*}[t]
    \centering
    \includegraphics[width=\textwidth, height=0.3\textheight, keepaspectratio]
    {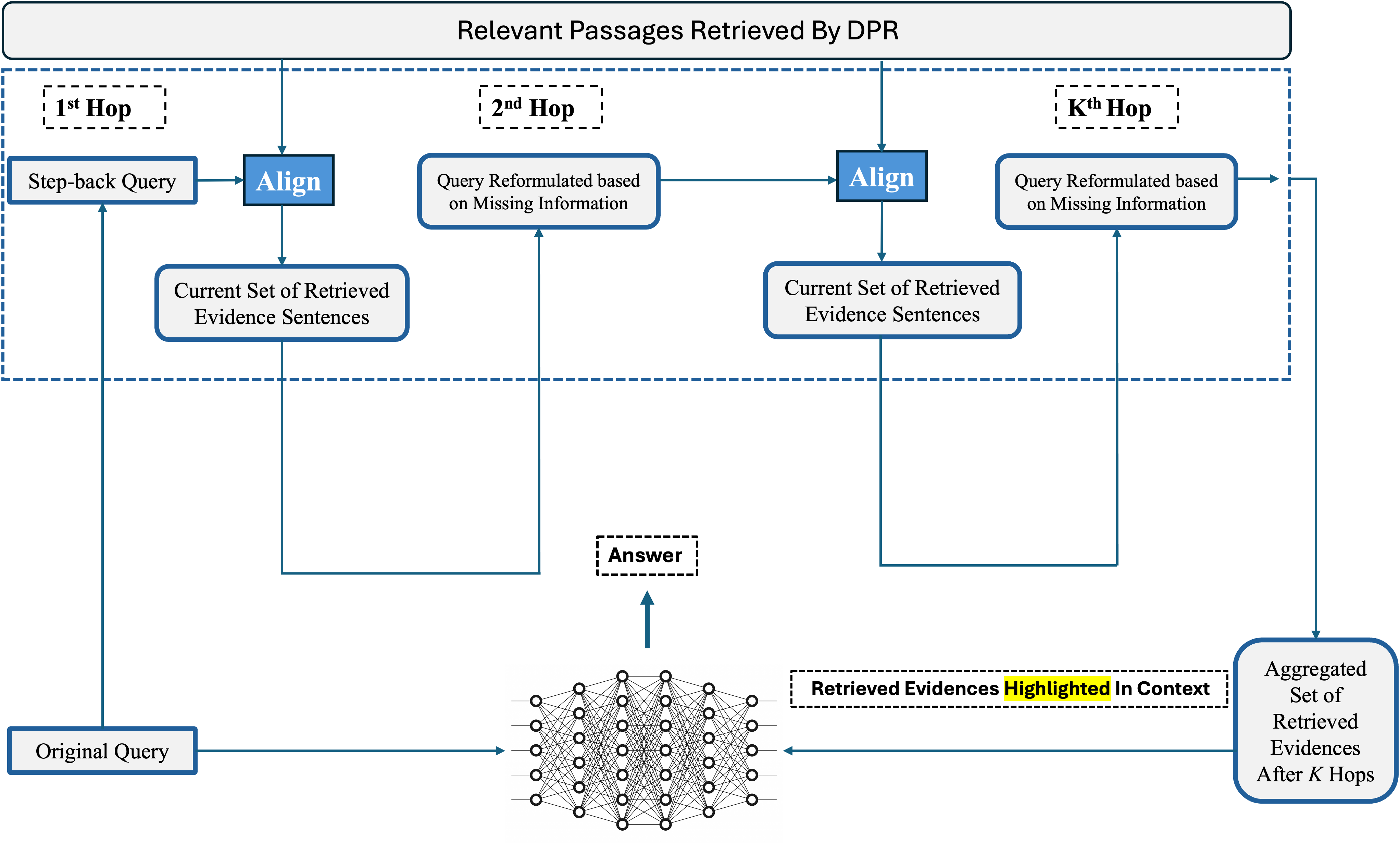}
    \caption{\footnotesize Our proposed method. Each query is concatenated with a more abstract \textit{Step-back} version of itself synthesized by a \textit{Step-back} LLM. This new query is used initiate multi-hop retrieval where in each hop the query is aligned with passages retrieved by DPR to select one evidence sentence. These sentences are aggregated across hops with alignment at each hop driven by query reformulation based on \textit{missing information} (maxim of relation) between the current set of selected evidence sentences and current query. After all query keywords are covered by the retrieved evidences (maxim of quantity), our method highlights them within their original contexts and provides them to the LLM.}
    \label{fig:SayLessRAG-diagram}
\end{figure*}

\begin{table*}[h]
  \centering
  \footnotesize
    \resizebox{\linewidth}{!}{
    \setlength{\tabcolsep}{3.5mm}{
    \begin{tabular}{llll}
    \toprule
    \textbf{\footnotesize Settings} & \textbf{\footnotesize ARC-C} & \textbf{\footnotesize PubHealth} & \textbf{\footnotesize PopQA} \\
    \midrule
    \textit{\footnotesize No Retrieval} \vspace{1mm} &       &       &  \\
    \footnotesize \quad Mistral-7B-Instruct \vspace{1mm} & \footnotesize 62.39 (\textcolor{darkgreen}{+6.72\%}) & \footnotesize 74.82 (\textcolor{darkgreen}{+0.96\%}) & \footnotesize 32.52 (\textcolor{red}{-49.73\%}) \\
    \footnotesize \quad Alpaca-7B \vspace{1mm} & \footnotesize 34.02 (\textcolor{red}{-17.43\%}) & \footnotesize 43.25 (\textcolor{red}{-7.78\%}) & \footnotesize 30.24 (\textcolor{red}{-53.04\%}) \\
    \footnotesize \quad Llama2-7B \vspace{1mm} & \footnotesize 40.94 (\textcolor{red}{-9.78\%}) & \footnotesize 68.02 (\textcolor{darkgreen}{+10.57\%}) & \footnotesize 23.73 (\textcolor{red}{-64.07\%}) \\
    \footnotesize \quad Qwen-2.5-3B \vspace{1mm} & \footnotesize \textbf{78.12} (\textcolor{darkgreen}{+7.28\%}) & \footnotesize 65.89 (\textcolor{red}{-7.15\%}) & \footnotesize 26.38 (\textcolor{red}{-62.39\%}) \\
    \footnotesize \quad AMD-OLMo-1B-SFT \vspace{1mm} & \footnotesize 25.81 (\textcolor{red}{-0.17\%}) & \footnotesize 60.81 (\textcolor{darkgreen}{+0.00\%}) & \footnotesize 33.38 (\textcolor{red}{-44.14\%}) \\
    \midrule
    \textit{\footnotesize DPR (No Evidence Highlighting)} \vspace{1mm} &       &       &  \\
    \footnotesize \quad Mistral-7B-Instruct \vspace{1mm} & \footnotesize 58.46 & \footnotesize 74.11 & \footnotesize 64.69 \\
    \footnotesize \quad Alpaca-7B \vspace{1mm} & \footnotesize 41.20 & \footnotesize 46.90 & \footnotesize 64.40 \\
    \footnotesize \quad Llama2-7B-chat \vspace{1mm} & \footnotesize 45.38 & \footnotesize 61.52 & \footnotesize 66.05 \\
    \footnotesize \quad Qwen-2.5-3B \vspace{1mm} & \footnotesize 73.33 & \footnotesize 70.96 & \footnotesize 75.48 \\
    \footnotesize \quad AMD-OLMo-1B-SFT \vspace{1mm} & \footnotesize 25.64 & \footnotesize 60.81 & \footnotesize 59.76 \\
    \midrule
    \textit{\footnotesize DPR + Evidence Highlighting + No Step-back} \vspace{1mm} &       &       &  \\
    \footnotesize \quad Mistral-7B-Instruct \vspace{1mm} & \footnotesize 59.23 (\textcolor{darkgreen}{+1.32\%}) & \footnotesize 76.04 (\textcolor{darkgreen}{+2.60\%}) & \footnotesize 63.90 (\textcolor{red}{-1.22\%}) \\
    \footnotesize \quad Alpaca-7B \vspace{1mm} & \footnotesize 41.28 (\textcolor{darkgreen}{+0.19\%}) & \footnotesize 50.56 (\textcolor{darkgreen}{+7.80\%}) & \footnotesize 63.83 (\textcolor{red}{-0.89\%}) \\
    \footnotesize \quad Llama2-7B-chat \vspace{1mm} & \footnotesize 47.44 (\textcolor{darkgreen}{+4.54\%}) & \footnotesize \footnotesize 62.64 (\textcolor{darkgreen}{+1.82\%}) & \footnotesize 65.98 (\textcolor{red}{-0.10\%}) \\
    \footnotesize \quad Qwen-2.5-3B \vspace{1mm} & \footnotesize 73.25 (\textcolor{red}{-0.11\%}) & \footnotesize 71.17 (\textcolor{darkgreen}{0.3\%}) & \footnotesize \textbf{77.34} (\textcolor{darkgreen}{+2.46\%}) \\
    \footnotesize \quad AMD-OLMo-1B-SFT \vspace{1mm} & \footnotesize 28.21 (\textcolor{darkgreen}{+10.02\%}) & \footnotesize 61.02 (\textcolor{darkgreen}{+0.35\%}) & \footnotesize 60.54 (\textcolor{darkgreen}{+1.31\%}) \\
    \midrule
    \textit{\footnotesize DPR + Evidence Highlighting + Step-back} \vspace{1mm} &       &       &  \\
    \footnotesize \quad Mistral-7B-Instruct & \footnotesize 59.57 (\textcolor{darkgreen}{+1.90\%}) & \footnotesize \textbf{76.14} (\textcolor{darkgreen}{+2.74\%}) & \footnotesize 64.19 (\textcolor{red}{-0.77\%}) \\
    \footnotesize \quad Alpaca-7B & \footnotesize 41.37 (\textcolor{darkgreen}{+0.41\%}) & \footnotesize 56.14 (\textcolor{darkgreen}{+19.70\%}) & \footnotesize 64.05 (\textcolor{red}{-0.54\%}) \\
    \footnotesize \quad Llama2-7B-chat \vspace{1mm} & \footnotesize 47.95 (\textcolor{darkgreen}{+5.66\%}) & \footnotesize 66.40 (\textcolor{darkgreen}{+7.94\%}) & \footnotesize 65.76 (\textcolor{red}{-0.43\%}) \\
    \footnotesize \quad Qwen-2.5-3B \vspace{1mm} & \footnotesize 73.08 (\textcolor{red}{-0.34\%}) & \footnotesize 70.15 (\textcolor{red}{-1.14\%}) & \footnotesize 77.48 (\textcolor{darkgreen}{+2.65\%}) \\
    \footnotesize \quad AMD-OLMo-1B-SFT \vspace{1mm} & \footnotesize 28.21 (\textcolor{darkgreen}{+10.02\%}) & \footnotesize 62.03 (\textcolor{darkgreen}{+2.01\%}) & \footnotesize 60.47 (\textcolor{darkgreen}{+1.19\%}) \\
    \bottomrule
    \end{tabular}%
    }
    }
    \caption{\footnotesize Our pragmatics driven RAG versus a Standard DPR RAG setup. \textbf{Bold} numbers indicate the best performance among all methods and LLMs for a specific dataset.  Percentage changes relative to the \textit{DPR without Evidence Highlighting} setting are shown in parentheses. Positive changes are highlighted in \textcolor{darkgreen}{green}, negative in \textcolor{red}{red}. In the \textit{No Retrieval} setting, we do not retrieve any documents and test the LLM's parametric knowledge. \textit{DPR (No Evidence Highlighting)} refers to the setting where we provide the top-$K$ passages for each query to the LLM without highlighting any evidence sentences within those passages. In the \textit{DPR + Evidence Highlighting + No Step-back} setting, we provide DPR passages annotated with highlighted evidences using ``$<$evidence$>$'' tokens. The \textit{DPR + Evidence Highlighting + Step-back} setting extends the previous setting by introducing reformulated queries and answer choices using Step-back prompting.
    }
  \label{tab:pragmatics_rag}%
\end{table*}
\begin{table*}[h]
\centering
\footnotesize 
\setlength{\tabcolsep}{3pt} 
\renewcommand{\arraystretch}{0.9} 
\begin{tabularx}{\textwidth}{>{\raggedright\arraybackslash}p{7cm} *{3}{>{\centering\arraybackslash}X}}
\toprule
\textbf{Dataset and Setting} & \textbf{Llama-2–7B-chat} & \textbf{Alpaca-7B} & \textbf{Mistral-7B-Instruct} \\
\midrule
ARC-C \textit{(Evidences w/ Context)} & 47.95 & 41.37 & 59.57 \\
ARC-C \textit{(Evidences w/o Context)} & 47.69 (\textcolor{red}{-0.54\%}) & 38.03 (\textcolor{red}{-8.07\%}) & 58.29 (\textcolor{red}{-2.14\%}) \\
PubHealth \textit{(Evidences w/ Context)} & 66.40 & 56.14 & 76.14 \\
PubHealth \textit{(Evidences w/o Context)} & 54.82 (\textcolor{red}{-17.44\%}) & 49.34 (\textcolor{red}{-12.11\%}) & 62.23 (\textcolor{red}{-18.27\%}) \\
\bottomrule
\end{tabularx}
\caption{\footnotesize Performance of various models on ARC-C and PubHealth datasets when using highlighted evidences within their original context versus using highlighted evidences while discarding surrounding context. Percentage changes (decreases) are shown in parentheses relative to the full context setting. Using highlighted evidence without its surrounding context can significantly degrade the LLMs QA performance. }
\label{tab:highlighted_justifications}
\end{table*}
\section{Approach: Combining Step-Back Reasoning With Pragmatic Retrieval}
\label{sec:approach}
Conceptually, our approach is a simple plug-and-play extension that emphasizes important information in any standard RAG setup (as shown in Figure \ref{fig:SayLessRAG-diagram}).
In this paper, we apply our extension to a collection of documents  retrieved by a dense passage retriever (DPR) \cite{izacard2021contriever}.\footnote{We use the same KB collection of documents as Self-RAG \cite{asai2024selfrag} and CRAG \cite{yan2024corrective}.} 
We adapt the unsupervised iterative sentence retriever proposed by \newcite{yadav2020unsupervisedalignmentbasediterativeevidence} to identify important sentences in the documents retrieved by RAG with DPR, as follows:
\textbf{(1)} Given a query and associated passages retrieved by DPR, the query is first conjoined with a more abstract \textit{step-back} version of itself created by a \textit{step-back LLM}~\cite{zheng2024stepbackevokingreasoning}. \textbf{(2)} In the first sentence retrieval iteration, this conjoined query is used to retrieve a set of relevant evidence sentences from the corresponding passages (see Eqs. 1 and 2). \textbf{(3)} In the next iteration(s), the query is reformulated to focus on \textit{missing information}, i.e., query keywords not covered by the current set of retrieved evidence sentences (see Eq. 3) and the process repeats until all question phrases are covered. As such, this strategy implements Grice's maxims of relation (because the evidence sentences are relevant to the question), quantity, and manner (because we identify as many sentences as needed to cover the question and no more).

By aggregating sets of retrieved evidence sentences across iterations, this retrieval strategy allows constructing \textit{chains} of evidence sentences for a given query, which can extend dynamically until a parameter-free termination criteria is reached. Further, by varying the first evidence sentence in the top $N$\footnote{In our experiments, we set $N=3$.} retrieved evidences, we can trivially extend this retriever to extract \textit{parallel evidence chains}, each of varying lengths, to create a more diverse set of evidence sentences that support the query. 

Lastly, we condition the generation of the Question Answering (QA) LLMs on the retrieved evidences, highlighted with special \textit{evidence tokens}, embedded in their original DPR contexts, in order (see Table~\ref{tab:example} for an example). 
We describe each of these stages in more detail below.

\subsection{Step-Back Query Expansion}
\label{sec:step-back}
In this work, we employ \textit{Step-Back Prompting} \cite{zheng2024stepbackevokingreasoning}, a simple technique to integrate LLM driven reasoning into the retrieval process. A step-back prompt elicits from the LLM an abstract, higher-level question derived from the original query, encouraging higher-level reasoning about the problem. For example, a step-back version of the query: ``As bank president, Alex Sink eliminated thousands of Florida jobs while taking over \$8 million in salary and bonuses. True or False?'' could be: ``What were the actions taken by Alex Sink as bank president?''. We hypothesize that step-back queries, representing a more generalized query formulation, when utilized as initialization seeds for the iterative retrieval (refer to Figure \ref{fig:SayLessRAG-diagram}), will generate a more diverse yet still relevant set of candidate evidence sentences. For multiple-choice questions (MCQs), we generate step-back answer choices for each option, combining them with the step-back query to guide retrieval. This approach introduces an additional dimension of parallelism in constructing evidence chains for MCQs. The step-back prompts used to elicit multi-hop reasoning follow the Knowledge QA template from \newcite{zheng2024stepbackevokingreasoning} (refer to appendix \ref{sec:appendix3} for prompts and Table \ref{tab:stepback-examples} for examples of step-back questions).
\subsection{Parallel Iterative Evidence Retrieval}
Computing an alignment score between queries and documents is a critical step in any retrieval system. Keeping in mind the Gricean maxim's of \textit{quality} and \textit{relation} (Section \ref{sec:intro}), which emphasize relevance and factual grounding, we leverage a principle similar to ``late interaction''
\citep{khattab2020colbertefficienteffectivepassage,santhanam2022colbertv2effectiveefficientretrieval}, where evidences are selected based on token-level similarities between queries and KB passages. We align query tokens with tokens from each sentence in the KB passages to construct evidence sentences, by selecting the most maximally similar token from the KB passage based on cosine similarity scores over dense embeddings\footnote{While \newcite{yadav2020unsupervisedalignmentbasediterativeevidence} align tokens based on similarity over GloVe embeddings, we use sentence transformer embeddings: \url{https://huggingface.co/jinaai/jina-embeddings-v2-base-en}} (Equation~\ref{eq:alignment_score}). 
\begin{equation}
\label{eq:alignment_score}
s(Q, P_j) = \sum_{i=1}^{|Q|} align(q_i, P_j)
\end{equation}

\begin{equation}
\label{eq:align_function}
align(q_i, P_j) = \max_{k=1}^{|P_j|} cosSim(q_i, p_k)
\end{equation}
where $q_i$ and $p_k$ are the $i^{th}$ and $k^{th}$ terms of the query $(Q)$ and evidence sentence $(P_j)$ respectively.

Query reformulation is driven by remainder terms, defined as the set of query terms which have not yet been covered by the set of evidence sentences which were retrieved in the first $i$ iterations of the multi-hop retriever (Equation~\ref{eq:remainder_terms}): 
\begin{equation}
\label{eq:remainder_terms}
Q_r(i) = t(Q) - \bigcup_{s_k \in S_i} t(s_k)
\end{equation}
where $t(Q)$ represents the unique set of query terms, $t(s_k)$ represents the unique terms of the $k^{th}$ evidence sentence in set $S_i$, which is the set of evidences retrieved in the $i^{th}$ iteration of the retrieval process.

The notion of coverage here is based on soft matching alignment: a query term is considered to be included in the set of evidence terms if its cosine similarity with a evidence term is greater than $M$\footnote{In this work, we set $M=0.98$.}. Note that the goal of query reformulation is to maximize the coverage of the query keywords by the retrieved chain of evidences, which aligns with the notion of the maxim of \textit{quantity} (Section \ref{sec:intro}).

Ambiguous queries are mitigated by dynamically expanding the current query with terms from all previously retrieved evidence sentences if the number of uncovered terms in the query falls below $T$,\footnote{In this work, we set $T=4$.} which also satisfies the last of Grice's maxims (maxim of \textit{manner}).
\begin{table*}[h]
  \centering
  \footnotesize
    \resizebox{\linewidth}{!}{
    \setlength{\tabcolsep}{3.5mm}{
    \begin{tabular}{llll}
    \toprule
    \textbf{\footnotesize Settings} & \textbf{\footnotesize ARC-C} & \textbf{\footnotesize PubHealth} & \textbf{\footnotesize PopQA} \\
    \midrule
    \textit{\footnotesize No Retrieval} \vspace{1mm} &       &       &  \\
    \footnotesize \quad Mistral-7B-Instruct \vspace{1mm} & \footnotesize 62.39 (\textcolor{darkgreen}{+9.11\%}) & \footnotesize \textbf{74.82} (\textcolor{darkgreen}{+34.23\%}) & \footnotesize 32.52 (\textcolor{darkgreen}{+18.17\%}) \\
    \footnotesize \quad Alpaca-7B \vspace{1mm} & \footnotesize 34.02 (\textcolor{red}{-16.02\%}) & \footnotesize 43.25 (\textcolor{darkgreen}{+17.05\%}) & \footnotesize 30.24 (\textcolor{red}{-22.66\%}) \\
    \footnotesize \quad Llama2-7B \vspace{1mm} & \footnotesize 40.94 (\textcolor{darkgreen}{+0.22\%}) & \footnotesize 68.02 (\textcolor{darkgreen}{+0.15\%}) & \footnotesize 23.73 (\textcolor{red}{-0.29\%}) \\
    \footnotesize \quad Qwen-2.5-3B \vspace{1mm} & \footnotesize \textbf{78.12} (\textcolor{darkgreen}{+6.30\%}) & \footnotesize 65.89 (\textcolor{darkgreen}{+51.30\%}) & \footnotesize 26.88 (\textcolor{darkgreen}{+0.83\%}) \\
    \footnotesize \quad AMD-OLMo-1B-SFT \vspace{1mm} & \footnotesize 25.81 (\textcolor{darkgreen}{+0.00\%}) & \footnotesize 60.81 (\textcolor{darkgreen}{+0.00\%}) & \footnotesize 33.38 (\textcolor{darkgreen}{+4.25\%}) \\
    \midrule
    \textit{\footnotesize BM25 (No Evidence Highlighting)} \vspace{1mm} &       &       &  \\
    \footnotesize \quad Mistral-7B-Instruct \vspace{1mm} & \footnotesize 57.18 & \footnotesize 55.74 & \footnotesize 27.52 \\
    \footnotesize \quad Alpaca-7B \vspace{1mm} & \footnotesize 40.51 & \footnotesize 36.95 & \footnotesize \textbf{39.10} \\
    \footnotesize \quad Llama2-7B \vspace{1mm} & \footnotesize 40.85 & \footnotesize 67.92 & \footnotesize 23.80 \\
    \footnotesize \quad Qwen-2.5-3B \vspace{1mm} & \footnotesize 73.50 & \footnotesize 43.55 & \footnotesize 32.38 \\
    \footnotesize \quad AMD-OLMo-1B-SFT  \vspace{1mm} & \footnotesize 25.81 & \footnotesize 60.81 & \footnotesize 32.02 \\
    \midrule
    \textit{\footnotesize BM25 + Evidence Highlighting + No Step-back} \vspace{1mm} &       &       &  \\
    \footnotesize \quad Mistral-7B-Instruct  \vspace{1mm} & \footnotesize 58.38 (\textcolor{darkgreen}{+2.10\%}) & \footnotesize 62.23 (\textcolor{darkgreen}{+11.64\%}) & \footnotesize 29.16 (\textcolor{darkgreen}{+5.96\%}) \\
    \footnotesize \quad Alpaca-7B \vspace{1mm} & \footnotesize 40.17 (\textcolor{red}{-0.84\%}) & \footnotesize 53.91 (\textcolor{darkgreen}{+45.90\%}) & \footnotesize 37.81 (\textcolor{red}{-3.30\%}) \\
    \footnotesize \quad Llama2-7B \vspace{1mm} & \footnotesize 47.69 (\textcolor{darkgreen}{+16.74\%}) & \footnotesize 62.23 (\textcolor{red}{-8.38\%}) & \footnotesize 33.88 (\textcolor{darkgreen}{+42.35\%}) \\
    \footnotesize \quad Qwen-2.5-3B \vspace{1mm} & \footnotesize 75.13 (\textcolor{darkgreen}{+2.22\%}) & \footnotesize 42.84 (\textcolor{red}{-1.63\%}) & \footnotesize 35.53 (\textcolor{darkgreen}{9.73\%}) \\
    \footnotesize \quad AMD-OLMo-1B-SFT  \vspace{1mm} & \footnotesize 25.13 (\textcolor{red}{-2.63\%}) & \footnotesize 59.39 (\textcolor{red}{-2.33\%}) & \footnotesize 33.10 (\textcolor{darkgreen}{+3.37\%}) \\
    \midrule
    \textit{\footnotesize BM25 + Evidence Highlighting + Step-back} \vspace{1mm} &       &       &  \\
    \footnotesize \quad Mistral-7B-Instruct \vspace{1mm} & \footnotesize 58.72 (\textcolor{darkgreen}{+2.69\%}) & \footnotesize 62.64 (\textcolor{darkgreen}{+12.38\%}) & \footnotesize 29.24 (\textcolor{darkgreen}{+6.25\%}) \\
    \footnotesize \quad Alpaca-7B \vspace{1mm} & \footnotesize 40.00 (\textcolor{red}{-1.26\%}) & \footnotesize 45.69 (\textcolor{darkgreen}{+23.65\%}) & \footnotesize 38.46 (\textcolor{red}{-1.64\%}) \\
    \footnotesize \quad Llama2-7B \vspace{1mm} & \footnotesize 47.61 (\textcolor{darkgreen}{+16.55\%}) & \footnotesize 61.93 (\textcolor{red}{-8.82\%}) & \footnotesize 34.31 (\textcolor{darkgreen}{+44.16\%}) \\
    \footnotesize \quad Qwen-2.5-3B  \vspace{1mm} & \footnotesize 74.62 (\textcolor{darkgreen}{+1.52\%}) & \footnotesize 43.05 (\textcolor{red}{-1.15\%}) & \footnotesize 34.88  (\textcolor{darkgreen}{7.72\%}) \\
    \footnotesize \quad AMD-OLMo-1B-SFT  \vspace{1mm} & \footnotesize 25.38 (\textcolor{red}{-1.67\%}) & \footnotesize 60.61 (\textcolor{red}{-0.33\%}) & \footnotesize 33.02 (\textcolor{darkgreen}{+3.12\%}) \\
    \bottomrule
    \end{tabular}%
    }
    }
    \caption{\footnotesize Our pragmatics driven RAG versus a BM25 RAG setup. \textbf{Bold} numbers indicate the best performance among all methods and LLMs for a specific dataset. Percentage changes relative to the BM25 \textit{without Evidence Highlighting} setting are shown in parentheses. Positive changes are highlighted in \textcolor{darkgreen}{green}, negative in \textcolor{red}{red}. In the \textit{No Retrieval} setting, we do not retrieve any documents and test the LLM's parametric knowledge. 
    \textit{BM25 (No Evidence Highlighting)} refers to the setting where we provide the top-$K$ passages for each query to the LLM without highlighting any evidence sentences within those passages.
    In the \textit{BM25 + Evidence Highlighting + No Step-back setting}, we provide BM25 passages annotated with highlighted evidences using ``$<$evidence$>$'' tokens.  The \textit{BM25 + Evidence Highlighting + Step-back} setting extends the previous setting by introducing reformulated queries and answer choices using Step-back prompting.}
  \label{tab:bm25_results}%
\end{table*}

\section{Results}
\label{sec:results}
\textbf{Evaluation \& Datasets} We evaluate our method on the test sets of ARC-Challenge (a \textit{MCQ} reasoning dataset), PubHealth (a fact \textit{verification} dataset about public health) \& PopQA (open-domain question-answering). For closed-tasks (ARC-Challenge, PubHealth), we evaluate Accuracy. For the short-form generation task (PopQA), the metrics indicate performance based on whether gold answers are included in the model generations instead of strictly requiring exact matching (Appendix \ref{sec:appendix4}). Table \ref{tab:pragmatics_rag} shows that integrating pragmatic hints into RAG can enhance performance over DPR. For example, on ARC-Challenge, combining evidence highlighting with step-back reasoning improves Llama-2-7B by up to 5.66\% and AMD-OLMo-1B by up to 10\% (relative, compared to using just the DPR passages without evidence highlighting). On PubHealth, our method improves Alpaca-7B by up to 19.7\% and Llama-2-7B by up to 7.94\%. In most cases, for both PubHealth and ARC-Challenge, the ``\textit{DPR + Evidence Highlighting + Step-back reasoning}'' setting consistently outperforms the ``\textit{Dense Passage Retrieval (DPR) (No Evidence Highlighting)}'' setting and the ``\textit{DPR + Evidence Highlighting + No Step-back reasoning}'' setting.\\
\textbf{Choice of LLMs} We primarily utilize older language models to mitigate data contamination risks \cite{sainz-etal-2023-nlp}. For instance, we excluded DeepSeek-R1-Distill-Llama-70B \cite{deepseekai2025deepseekr1incentivizingreasoningcapability} after observing its 90\% accuracy on ARC-Challenge under \textit{No Retrieval Setting}---a clear indication of data leakage. While our selected models may still exhibit some contamination (evidenced by strong performance in \textit{No Retrieval} settings), our method demonstrates improvements over these models even when paired with Dense Passage Retrieval, establishing a comparative baseline. Please refer to Appendix \ref{sec:appendix4} for details of the prompts used and other experimental details.

\begin{figure}[!htb]
  \centering
    \includegraphics[width=\columnwidth]{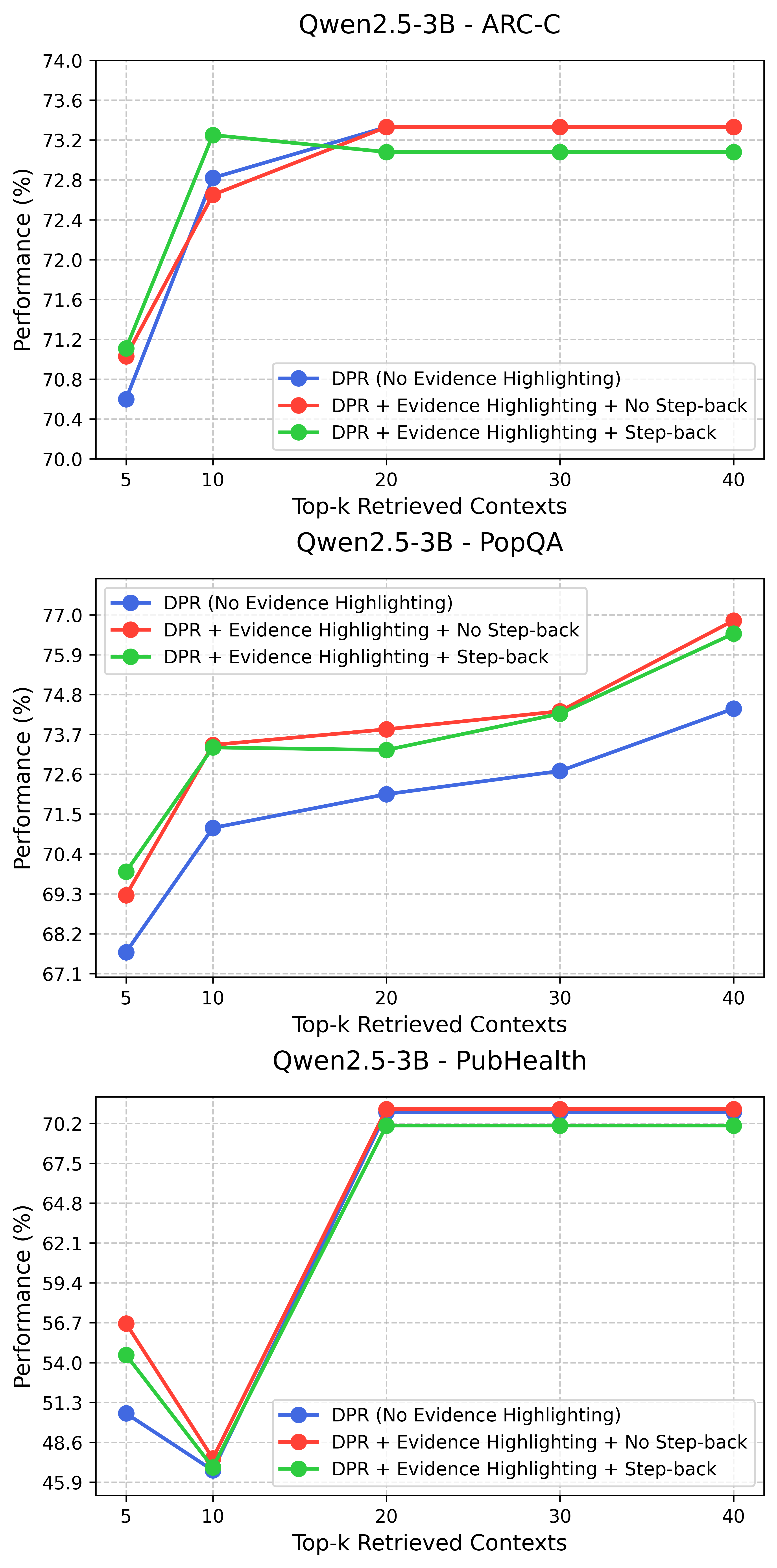}
  \caption{Performance of Qwen2.5-3B on: (\textbf{top}) ARC-Challenge, \textbf{(middle)} PopQA, (\textbf{bottom}) PubHealth under different \textit{Evidence Highlighting} settings, with varying top-$k$ where $k$ is the number of DPR contexts retrieved.}
  \label{fig:top-k}
\end{figure}


\begin{table*}[h]
\centering
\footnotesize
\begin{tabular}{lcc}
\toprule
\textbf{Category} & \textbf{Frequency (ARC-Challenge)} & \textbf{Frequency (PubHealth)} \\
\midrule
\textbf{Bad} (0) & 6 & 8 \\
\textbf{Medium} (0.5) & 10 & 4 \\
\textbf{Good} (1) & 4 & 8 \\
\bottomrule \\
\end{tabular}
\caption{\footnotesize Highlighted Evidence Quality Scores for 20 randomly sampled queries from the ARC-Challenge and PubHealth datasets. The frequencies represent the number of instances falling into each quality category for the highlighted evidence in both datasets.}
\label{tab:evidence_quality_scores}
\end{table*}

\section{Analysis}
\textbf{When Does Pragmatics Help?}
Our error analysis indicates that leveraging pragmatics is effective when answering the query requires connecting facts along a causal path to deduce the answer (as shown in the example of Good Evidence in Table \ref{tab:evidence_quality_examples}, appendix \ref{sec:appendix1}). 

We also observe that highlighted evidence often functions as implicit few-shot exemplars, facilitating analogical reasoning. For instance, given the question ``In the design process, what is an example of a trade-off?'', our method highlights two analogous scenarios: a career decision (``\$50,000 salary worker sacrificing income to pursue medical training with the goal of increasing their future income after becoming a doctor'') and a biological principle (``beneficial trait changes linked to detrimental ones''). We hypothesize that such examples stimulate the model's in-context learning capabilities, possibly explaining the observed 10\% relative improvement in OLMo-1B's performance on ARC-C.

However, our method exhibits a few limitations in specific scenarios (refer to Table \ref{tab:low_quality_evidences}, appendix \ref{sec:appendix2}). First, it fails to highlight relevant evidences for queries which require arithmetic manipulation or comparison of physical quantities, as these tasks depend more on mathematical reasoning than factual knowledge. Second, it struggles with complex linguistic phenomena, particularly negation patterns. For example, consider the question: ``Which human activities would have a positive effect on the natural environment?'' Most retrieved passages focus on negative environmental impacts, reflecting their prevalence in real world corpora. The task here requires identifying contrary evidence from the long tail of the distribution, but our unsupervised retrieval heuristics do not account for such semantic inversions. \\
Lastly, we find that for factoid QA tasks like PopQA, evidence highlighting can slightly degrade performance compared to DPR, likely because these tasks rely more on the model's parametric knowledge. For instance, PopQA queries like ``What is Antonio Álvarez Alonso's occupation?'' often retrieve ambiguous contexts with multiple roles (e.g., Spanish retired footballer, Spanish paracanoeist, Spanish pianist and composer), offering insufficient signals for disambiguation. In such scenarios, our method may either highlight all potential evidences or arbitrarily select one, confusing the model and potentially leading to incorrect answers.

\textbf{Time Complexity of Retrieval}
The computational complexity of our retrieval method can be decomposed into two main components: First, for every query, we make one call to a step-back LLM for query expansion (i.e., creating an abstract step-back version of the query, refer to section \ref{sec:step-back}). Second, for evidence selection and highlighting \cite{yadav2020unsupervisedalignmentbasediterativeevidence}, given $S$ sentences retrieved by DPR, we select a subset of $K$ evidence sentences  from $S$ passage sentences. In each hop of the iterative retriever, one evidence sentence is chosen from $S$. The number of hops is upper bounded by the hyperparameter $K$ (where we set $K \leq 6$). 
Thus the cost of this step is $O(K \times S)$ (constant). Since we allow the retriever to extract $N$ parallel evidence chains by varying the top-scoring evidence (see section \ref{sec:approach}), the total cost of parallel evidence retrieval is $O(N \times K \times S)$ (constant). Evidence highlighting requires a linear scan of the $S$ passage sentences with complexity $O(S)$ (constant). Therefore, the total computational complexity is: $\text{Cost}_{\text{total}} = \text{Cost}({\text{LLM}_{\text{stepback}}}) + O(n)$, where $n$ represents the number of tokens in the retrieved passages $S$. We note two important considerations: (a) the base retrieval cost is inherent to any RAG system and thus unavoidable, and (b) our method introduces minimal computational overhead compared to alternative reasoning-enhanced QA approaches such as STaR \cite{zelikman2022starbootstrappingreasoningreasoning}.

\textbf{Is keeping full DPR context necessary?}
We conduct an experiment to assess the impact of removing versus retaining the surrounding context of highlighted evidence sentences on QA performance. As shown in Table \ref{tab:highlighted_justifications}, across both ARC-C and PubHealth datasets and three different LLMs, we find that providing only the highlighted evidence sentences—without their surrounding context—can significantly degrade QA performance compared to retaining the full context while highlighting the evidence within.

\textbf{How does the quality of the retrieved passages impact our method}? To assess the relationship between initial retrieval quality and our method's effectiveness, we conduct comparative experiments using the sparse retrieval method BM25 \cite{Robertson2009ThePR} in place of DPR. For each query, we retrieve the top-20 passages using BM25, then apply our iterative retrieval approach with step-back reasoning (Section \ref{sec:approach}) to identify and highlight key evidence sentences within these contexts. As shown in Table \ref{tab:bm25_results}, retrieval quality significantly influences our method's performance. We observe substantial improvements across multiple models and datasets: Llama-2-7B achieves a 16.74\% gain on ARC-Challenge, Alpaca-7B shows up to a 45.90\% improvement on PubHealth, while Llama-7B and Qwen2.5-3B demonstrate gains of up to 44.16\% and 9.73\% on PopQA, respectively, relative to their baseline BM25 performance. However, the efficacy of our method when applied to BM25-retrieved passages is inconsistent, with several models also demonstrating performance deterioration compared to both baseline BM25 and the ``\textit{No Retrieval}'' setting. We hypothesize that this is because of two reasons: (a) 
BM25's lexical overlap-based retrieval mechanism yields passages containing necessary but insufficient information for query resolution.
For instance, on ARC-Challenge (refer to Table \ref{tab:bm25_results}), Alpaca-7B improves by 16\% when using BM25-retrieved passages as context, but subsequent evidence highlighting on top of these passages diminishes this gain. (b) Evidence highlighting more effectively grounds the LLM in the retrieved context, potentially overriding useful parametric knowledge. This effect is particularly pronounced with Qwen-2.5 3B on PubHealth, where the model significantly degrades by 51.3\% when provided with BM25 retrieved passages as contexts relative to ``\textit{No Retrieval}'',  and the application of evidence highlighting over these contexts further reduces performance by 1.6\%. This suggests that while evidence highlighting effectively directs model attention in high-quality passages, it creates a bias that may be counterproductive when retrieved passages are of lower quality. \footnote{We do not imply that BM25-retrieved passages are always lower quality than those retrieved by DPR; rather, in this specific case, the DPR \textit{Contriever} has been finetuned on web-domain data \cite{bajaj2018msmarcohumangenerated} similar to our evaluation datasets, making it a more effective retrieval method. We acknowledge that BM25 can be more robust than DPR out-of-domain.}
In such instances, our method may constrain the model to prioritize highlighted information over potentially superior parametric knowledge (which the model may have acquired through test data appearing in its pre-training corpora). These results suggest that our approach is more complementary to DPR and similar neural retrieval methods than to lexical matching approaches like BM25.

\textbf{Evaluating Quality of Highlighted Evidence}
We conduct a human evaluation to assess the quality of evidence highlighting across 40 questions, evenly distributed between the ARC-Challenge and PubHealth datasets (20 questions from each). For each question, we use a three-point rating scale to evaluate the corresponding highlighted evidence: \textbf{0 (bad)}, \textbf{0.5 (medium)} and \textbf{1 (good)}. Overall, 60\% to 70\%  of highlighted evidences were rated at least ``medium'' by the human evaluator across both datasets. See Appendix \ref{sec:appendix1} for the evaluation criteria used and examples of `good', `medium' and `bad' evidence sentences.

\textbf{Understanding the Impact of Top-\textit{k} Retrieval}
We analyze the effect of varying DPR's top-$k$ retrieved contexts on Qwen2.5-3B's performance in each of our three settings: vanilla DPR, DPR with evidence highlighting, DPR with evidence highlighting and step-back reasoning. Our results (figure \ref{fig:top-k}) indicate that larger $k$ values generally improve performance on each dataset. On ARC-C, we see a ``Goldilocks zone'' with step-back reasoning: increasing context ($k>10$) degrades performance. On PopQA, both evidence highlighting methods outperform vanilla DPR for all $k$ values tested. On PubHealth, we observe that evidence highlighting methods can significantly outperform DPR for smaller values of $k$ ($<10$).

\section{Conclusions}
We present an unsupervised method that enhances retrieval-augmented generation (RAG) by highlighting key sentences in retrieved documents. We find that this approach can improve QA performance across 3 different datasets and 5 different LLMs.

\section*{Limitations}
This study investigates the effectiveness of pragmatics in enhancing Retrieval Augmented Generation (RAG) systems. Our evaluation, however, is limited to a comparison against standard Dense Passage Retriever (DPR) and BM25 baselines. The proposed method has potential for integration with more sophisticated RAG systems, such as those developed by \newcite{asai2024selfrag, xu2024recomp, sarthi2024raptor}. Our assessment encompasses three datasets, but a more comprehensive evaluation would involve a broader range of single-hop and multi-hop tasks. Moreover, there are several scenarios which our approach does not cover, such as handling linguistic phenomena like negation, mathematical reasoning tasks and reconciling retrieved contexts that are ambiguous. Our current approach is also limited by the fact that it is unsupervised and query reformulation is mostly driven by a bag-of-words. One could trivially improve query reformulation by using an LLM, or using a weakly supervised strategy that fine-tunes an LLM to retrieve pragmatic evidence (using supervision from the current retriever) via a joint loss that learns to retrieve evidence sentences while simultaneously answering the query correctly (motivated by the relevance estimator and answer marginalization losses proposed by \newcite{kim2024reragimprovingopendomainqa}). We leave the exploration of supervised pragmatic RAG methods as future work.

While we hypothesize that our retrieved \& highlighted justifications constitute ``shallow chains of thought'' which are faithfully utilized by the Large Language Model in its generations, this assertion remains to be formally validated through rigorous analysis.
\bibliography{custom}
\appendix

\begin{table*}[t]
\centering
\footnotesize
\begin{tabularx}{\textwidth}{lX}
\toprule
\textbf{Category} & \textbf{Examples of Evidences} \\
\midrule
Good Evidence &
\textbf{Question:} A certain atom has 20 electrons, 21 neutrons, and 20 protons. What is the atomic mass of the atom?\\[1ex]
& \textbf{Highlighted Evidence}:\\[0.5ex]
& \quad $-$ ``Mass number (symbol `A', from German `Atomgewicht') is the total number of protons and neutrons (nucleons) in a nucleus.''\\[0.5ex]
& \quad $-$ ``Atomic mass is approximately the mass number times an atomic mass unit (approximate mass of a proton, neutron, or hydrogen-1 atom).''\\[1ex]
Medium Evidence &
\textbf{Question:} A law in Japan makes it illegal for citizens of that country to be fat.\\[1ex]
& \textbf{Highlighted Evidence}:\\[0.5ex]
& \quad $-$ ``Japan implemented the `metabo' law in 2008 to combat rising obesity rates.''\\[0.5ex]
& \quad $-$ ``The New York Times reported that the law aims to shrink the overweight population by 10\% over 4 years and 25\% over 7 years via financial penalties.''\\[0.5ex]
& \quad $-$ ``In 2008, Japan passed the ``Metabo Law,'' addressing metabolic syndrome—a cluster of conditions increasing the risk of heart disease, stroke, and diabetes.''\\[0.5ex]
& \quad $-$ ``The law requires models to have a minimum BMI and warns against photoshopped images.''\\[1ex]
Bad Evidence &
\textbf{Question:} Ted Cruz Says Democrats are embracing abortion up until (and even after) birth.\\[1ex]
& \textbf{Highlighted Evidence}:\\[0.5ex]
& \quad $-$ ``In January 2016, Cruz announced his "Pro-Lifers for Cruz" coalition, with statements about executing abortion doctors to expunge bloodguilt.''\\[0.5ex]
& \quad $-$ ``Kamala Harris refuted Republican claims about Democrats' abortion views.''\\[0.5ex]
& \quad $-$ ``In the mid-1990s, Moynihan supported banning the procedure known as partial-birth abortion.''\\
\bottomrule
\end{tabularx}
\caption{Examples of Good, Medium, and Bad Highlighted Evidences}
\label{tab:evidence_quality_examples}
\end{table*}

\begin{table*}[t]
\centering
\footnotesize
\begin{tabularx}{\textwidth}{lX}
\toprule
\midrule
ARC-Challenge & 
\textbf{Question:} Scott filled a tray with juice and put it in a freezer. The next day, Scott opened the freezer. How did the juice most likely change?\\[1ex]
& \textbf{Evidence:}\\
& \quad - Most recently, Scott produced the documentary film “Apple Pushers” with Joe Cross (filmmaker) juicer and a generator.\\[0.5ex]
& \quad - However, in March 1996, 70,000 Juice Tiger juicers (9\% of its models) were recalled after 14 injury incidents were reported.\\[1ex]
ARC-Challenge & 
\textbf{Question:} A physicist wants to determine the speed a car must reach to jump over a ramp. The physicist conducts three trials. In trials two and three, the speed of the car is increased by 20 miles per hour. What is the physicist investigating when he changes the speed?\\[1ex]
& \textbf{Evidence:}\\
& \quad - Objects in motion often have variations in speed (a car might travel at 50 km/h, slow to 0 km/h, then reach 30 km/h).\\[0.5ex]
& \quad - Preparing an object for g-tolerance (avoiding damage when subjected to high speeds).\\[0.5ex]
& \quad - Hence, the round-trip time on traveler clocks will be $\Delta \tau = 4\left(\frac{c}{\alpha}\right)\cosh(\gamma)$.\\[1ex]
ARC-Challenge & 
\textbf{Question:} Human activities affect the natural environment in many ways. Which action would have a positive effect on the natural environment?\\[1ex]
& \textbf{Evidence:}\\
& \quad - This environment encompasses the interaction of all living species, climate, weather, and natural resources affecting human survival and economic activity.\\[0.5ex]
& \quad - For instance, actions by the U.S. Army Corps of Engineers that threatened ecosystems in Florida's Oklawaha River valley and issues in preserving Pacific Coast Redwood communities are cited as case studies.\\[0.5ex]
& \quad - Humans have contributed to the extinction of many plants and animals.\\
Pop-QA & 
\textbf{Question:} What is Antonio Álvarez Alonso's occupation?\\[1ex]
& \textbf{Evidence:}\\
& \quad - Antonio De Diego  Antonio de Diego Álvarez is a Spanish paracanoeist and member of the National Spanish Canoeist Team, Paracanoe class A (maximum level of disability).\\[0.5ex]
& \quad - Antonio Álvarez Alonso  Antonio Álvarez Alonso (11 March 1867 - 22 June 1903) was a Spanish pianist and composer.\\[0.5ex]
\bottomrule
\end{tabularx}
\caption{Examples of low-quality evidences retrieved for various types of queries from ARC-Challenge \& Pop-QA}
\label{tab:low_quality_evidences}
\end{table*}

\begin{table*}[t]
\centering
\footnotesize
\begin{tabularx}{\textwidth}{lX}
\toprule
\textbf{Dataset} & \textbf{Original Question and Step-back Question} \\
\midrule
ARC-Challenge &
\textbf{Original Question:} An astronomer observes that a planet rotates faster after a meteorite impact. Which is the most likely effect of this increase in rotation?\\[1ex]
& \textbf{Step-back Question:} What effects do meteorite impacts on planets have?\\[1ex]
ARC-Challenge &
\textbf{Original Question:} A group of engineers wanted to know how different building designs would respond during an earthquake. They made several models of buildings and tested each for its ability to withstand earthquake conditions. Which will most likely result from testing different building designs?\\[1ex]
& \textbf{Step-back Question:} What are the testing methods used by the engineers to determine the earthquake resilience of the different building models?\\[1ex]
PopQA &
\textbf{Original Question:} What is Henry Feilden's occupation?\\[1ex]
& \textbf{Step-back Question:} What are the important aspects of Henry Feilden's academic work?\\[1ex]
PubHealth &
\textbf{Original Question:} A mother revealed to her child in a letter after her death that she had just one eye because she had donated the other to him.\\[1ex]
& \textbf{Step-back Question:} What are the circumstances surrounding the donation of the mother's second eye to her child after her death?\\
\bottomrule
\end{tabularx}
\caption{Examples of Step-back questions created from original questions in the three datasets.}
\label{tab:stepback-examples}
\end{table*}

\section{Human Evaluation of Evidence Quality}
\label{sec:appendix1}
\subsection{Evaluation Criteria}
For each query, we categorize its corresponding set of highlighted evidence(s) as ``bad'' (score: 0) when it includes completely irrelevant sentences or sentences within contexts that are somewhat related to the query but fail to provide any meaningful support in addressing it. In the case of fact-checking datasets like PubHealth, we also classify highlighted evidence as ``bad'' if it appears to support a claim but overlooks negations in the surrounding context that would ultimately refute the claim.

Highlighted evidence is categorized as ``medium'' (score: 0.5) when it consists of sentences situated in relevant contexts that may allow the correct answer to be inferred indirectly in some instances but lack the direct or explicit support needed to answer the query.

Highlighted evidence is categorized as ``good'' (score: 1) when it includes a sufficient number of sentences that directly address the query while ensuring no confounding factors (e.g., negations in the surrounding context) are overlooked.

Table \ref{tab:evidence_quality_examples} shows an example of good, medium and bad quality evidences as assessed by a human evaluator. The example of \texttt{Good Evidence} shown is rated as such because connecting the evidence sentences together allows the reader to deduce the answer to the query ``What is the atomic mass of the atom?'' even without extensive prior knowledge of chemistry.

\section{Low Quality Evidence}
\label{sec:appendix2}
In Table \ref{tab:low_quality_evidences}, we include some examples of retrieved evidences from the ARC-C dataset that do not help the model to deal with specific tasks, especially those which requiring modeling negation and arithmetic reasoning.

\section{Step-Back Reasoning Examples}
\label{sec:appendix3}
Please refer to Table \ref{tab:stepback-examples} for examples of original queries and the more abstract \textit{Step-back} questions elicited from those queries.

\subsection{Step-back Prompt for Query Expansion}
\begin{lstlisting}
You are an expert at world knowledge. Your task is to step back and paraphrase a question to a more generic step-back question, which is easier to answer. Here are a few examples:

Original Question: Which position did Knox Cunningham hold from May 1955 to Apr 1956?
Stepback Question: Which positions have Knox Cunningham held in his career?
        
Original Question: who has scored most runs in t20 matches as of 2017
Stepback Question: What are the runs of players in t20 matches as of 2017
        
Original Question: When was the abolishment of the studio that distributed The Game?
Stepback Question: which studio distributed The Game?
        
Original Question: What city is the person who broadened the doctrine of philosophy of language from?
Stepback Question: who broadened the doctrine of philosophy of language
        
Original Question: Would a Monoamine Oxidase candy bar cheer up a depressed friend?
Stepback Question: What are the effects of Monoamine Oxidase?
        
What is the Stepback Question for this?: {original_question_text}
Answer with only the Stepback Question and no extra text.
\end{lstlisting}

\subsection{Step-back Prompt for MCQ Answer Choices}
\begin{lstlisting}
You are an expert at world knowledge. You are given a statement. Your task is to extract the concepts and principles underlying the statement. Answer only with the concepts and principles without any extra text.
If there are multiple concepts and principles, list them separated by commas.
Original Statement: {answer_text}
Answer:\end{lstlisting}

\section{Experimental Details}
\label{sec:appendix4}
Our experimental results for Mistral-7B-Instruct v0.1, Alpaca-7B \& Llama-2-7B differ from those reported by other works such as Self-RAG \cite{asai2024selfrag} \& CRAG \cite{yan2024corrective}, and Speculative RAG due to the following methodological variations:
\begin{enumerate}
    \item \textbf{Evaluation Function:} We employ a different evaluation criteria for assessing accuracy between Large Language Model (LLM) generations and gold labels in tasks such as ARC-Challenge, PopQA, and PubQA. Our approach considers an LLM generation correct based on the principle of ``inclusion,'' i.e., if the generation includes the correct answer as a substring, post-normalization.
    
    \item \textbf{Number of retrieved passages in DPR and BM25 (top-K):} In both BM25 and DPR retrieval, we set $K=11$ for models which have a 4096 token limit context (e.g., Llama-2-7B), where 10 passages are from the Wikipedia KB mixed with a web search result from CRAG.
    For Alpaca-7B and AMD-OLMo-1B-SFT, owing to their small context window size of 2048, we keep just the top-9 documents ($K=9$). For Alpaca and OlMo, we observe significant degradation if we use 10 or more documents causing the DPR setting to perform worse than even the \textit{No-Retrieval model}. For models with larger context windows e.g., Mistral-7B and Qwen2.5-3B we use all DPR and BM25 retrieved passages.

    \item \textbf{Prompt Engineering:} Our prompts differ slightly from those used in Self-RAG and C-RAG. We have engineered our prompts to adhere more closely to the recommended Instruction Tuning format, particularly for Alpaca-7B \cite{alpaca} and Llama-2-7B-chat \cite{touvron2023llama2openfoundation}.
    \item \textbf{Stepback-LLM:} In all experiments, we use Mistral-7B-Instruct v0.1 as the step-back LLM.
\end{enumerate}
\noindent These methodological distinctions should be considered when comparing our results with those of previous studies.

\section{Example Prompts}
\label{sec:appendix5}
Examples of the task specific prompts utilized in our study are as follows:
\small
\begin{itemize}
    \item \textbf{ARC-Challenge}
    \begin{itemize}
        \item Mistral-7B-Instruct:
        \begin{lstlisting}[basicstyle=\small\ttfamily,breaklines=true]
Refer to the following documents, follow the instruction and answer the question.

Documents: {highlighted_passages}

Question: {question}

Instruction: Given four answer candidates, A, B, C and D, choose the best answer choice.
Please answer with the capitalized alphabet only, without adding any extra phrase or period.
        \end{lstlisting}
        
        \item Alpaca-7B:
        \begin{lstlisting}[basicstyle=\small\ttfamily,breaklines=true]
Below is an instruction that describes a task. Write a response that appropriately completes
the request.

### Instruction: Given four answer candidates, A, B, C and D, choose the best answer choice.
Please answer with the capitalized alphabet only, without adding any extra phrase or period.

### Input:
Documents: {highlighted_passages}
Question: {question}
Choices: {choices_str}

### Response: 
        \end{lstlisting}

        \item Llama-2-7B-chat:
        \begin{lstlisting}[basicstyle=\small\ttfamily,breaklines=true]
Below is an instruction that describes a task. Write a response that appropriately completes
the request.

### Instruction: Given four answer candidates, A, B, C and D, choose the best answer choice.
Please answer with the capitalized alphabet only, without adding any extra phrase or period.

### Input:
Documents: {highlighted_passages}
Question: {question}
Choices: {choices_str}

### Response:  
        \end{lstlisting}
    \end{itemize}

    \item \textbf{PopQA}
    \begin{itemize}
        \item Mistral-7B-Instruct:
        \begin{lstlisting}[basicstyle=\small\ttfamily,breaklines=true]
Refer to the following documents, follow the instruction and answer the question.

### Input:
Documents: {highlighted_passages}

### Instruction: Answer the question: {question}
### Response:
        \end{lstlisting}        
        \item Alpaca-7B:
        \begin{lstlisting}[basicstyle=\small\ttfamily,breaklines=true]
Below is an instruction that describes a task. Write a response that appropriately completes
the request.

### Instruction: Refer to the following documents and answer the question.
### Input:
Documents: {highlighted_passages}

Question: {question}
### Response: 
        \end{lstlisting}
        \item Llama-2-7B:
        \begin{lstlisting}[basicstyle=\small\ttfamily,breaklines=true]
<s>[INST] <<SYS>>
    You are a helpful, respectful and honest assistant. Always answer as helpfully as possible,
    while being safe. Your answers should not include any harmful, unethical, racist, sexist,
    toxic, dangerous, or illegal content. Please ensure that your responses are socially unbiased
    and positive in nature.

    If a question does not make any sense, or is not factually coherent, explain why instead of
    answering something not correct. If you don't know the answer to a question, please don't
    share false information.
<</SYS>>

Below is an instruction that describes a task. Write a response that appropriately completes
the request.

Instruction: Refer to the following documents and answer the question.

Documents: {highlighted_passages}

Question: {question}
### Response: [/INST]
        \end{lstlisting}
    \end{itemize}
    
    \item \textbf{PubHealth}
    \begin{itemize}
        \item Mistral-7B-Instruct:
        \begin{lstlisting}[basicstyle=\small\ttfamily,breaklines=true]
Read the documents and answer the question: Is the following statement correct or not?
Only say true if the statement is true; otherwise say false. Don't capitalize or add periods,
just say ``true'' or ``false''.

Documents: {highlighted_passages}

Statement: {question}
### Response:
        \end{lstlisting}
        \item Alpaca-7B:
        \begin{lstlisting}[basicstyle=\small\ttfamily,breaklines=true]
Below is an instruction that describes a task. Write a response that appropriately completes
the request.

### Instruction: Read the documents and answer the question: Is the following statement correct
or not? Only say true if the statement is true; otherwise say false. Don't capitalize or add
periods, just say ``true'' or ``false''.

### Input:
Documents: {highlighted_passages}

Statement: {question}
### Response: 
        \end{lstlisting}
        \item Llama-2-7B:
        \begin{lstlisting}[basicstyle=\small\ttfamily,breaklines=true]
<s>[INST] <<SYS>>
    You are a helpful, respectful and honest assistant. Always answer as helpfully as possible,
    while being safe. Your answers should not include any harmful, unethical, racist, sexist,
    toxic, dangerous, or illegal content. Please ensure that your responses are socially unbiased
    and positive in nature.

    If a question does not make any sense, or is not factually coherent, explain why instead of
    answering something not correct. If you don't know the answer to a question, please don't
    share false information.
<</SYS>>

Below is an instruction that describes a task. Write a response that appropriately completes
the request.

### Instruction: Read the documents and answer the question: Is the following statement correct or not? Only say true if the statement is true; otherwise say false. Don't capitalize or add
periods, just say ``true'' or ``false''.

### Input:
Documents: {highlighted_passages}

Statement: {question}
### Response: [/INST]
        \end{lstlisting}
    \end{itemize}
\end{itemize}
\normalsize

\end{document}